\documentclass{bmvc2k}

\usepackage{times}
\usepackage{epsfig}
\usepackage{graphicx}
\usepackage{amsmath}
\usepackage{amssymb}
\usepackage{comment}
\usepackage{bm}
\usepackage{here}

\title{Intentional Attention Mask Transformation for Robust CNN Classification}

\addauthor{Masanari Kimura}{}{1}
\addauthor{Masayuki Tanaka}{}{1}

\addinstitution{
 National Institute of Advanced Industrial Science and Technology,\\
 Tokyo, Japan
}
\addinstitution{
 Tokyo Institute of Technology
 Tokyo, Japan
}

\runninghead{Student, Prof, Collaborator}{Intentional Attention Mask Transformation}

\begin{document}

\maketitle

\begin{abstract}
Convolutional Neural Networks have achieved impressive results in various tasks, but interpreting the internal mechanism is a challenging problem. To tackle this problem, we exploit a multi-channel attention mechanism in feature space. Our network architecture allows us to obtain an attention mask for each feature while existing CNN visualization methods provide only a common attention mask for all features.
We apply the proposed multi-channel attention mechanism to multi-attribute recognition task. We can obtain different attention mask for each feature and for each attribute. Those analyses give us deeper insight into the feature space of CNNs.
Furthermore, our proposed attention mechanism naturally derives a method for improving the robustness of CNNs.
From the observation of feature space based on the proposed attention mask, we demonstrate that we can obtain robust CNNs by intentionally emphasizing features that are important for attributes.
The experimental results for the benchmark dataset show that the proposed method gives high human interpretability while accurately grasping the attributes of the data, and improves network robustness.

\end{abstract}

\section{Introduction}
In recent years, Convolutional Neural Networks (CNNs) have made great achievements in various tasks \cite{krizhevsky2012imagenet, karpathy2014large}.
Despite such success, it is known that an interpretation of the CNNs is difficult for humans.
Therefore, visual explanation, which 
visualizes the inference mechanism of the CNNs, is becoming one of hot topics~\cite{selvaraju2017grad, zhang2018visual, kuwajima2019improving, kimura2019interpretation}.
In this paper, we introduce a multi-channel attention sub-networks to improve interpretability of CNNs.
Our main idea is to train sub-networks with multi-channel attention mask for each attribute.
The multi-channel attention sub-networks provide us feature-dependent attention masks, while existing attention mask approach~\cite{fukui2018attention, wang2017residual, hu2018squeeze} only provides common single attention mask.
This multi-channel attention mechanism can reveal which channel in the feature map focuses on which part of the image.

\begin{figure*}[tb]
  \centering
  \includegraphics[scale=0.32]{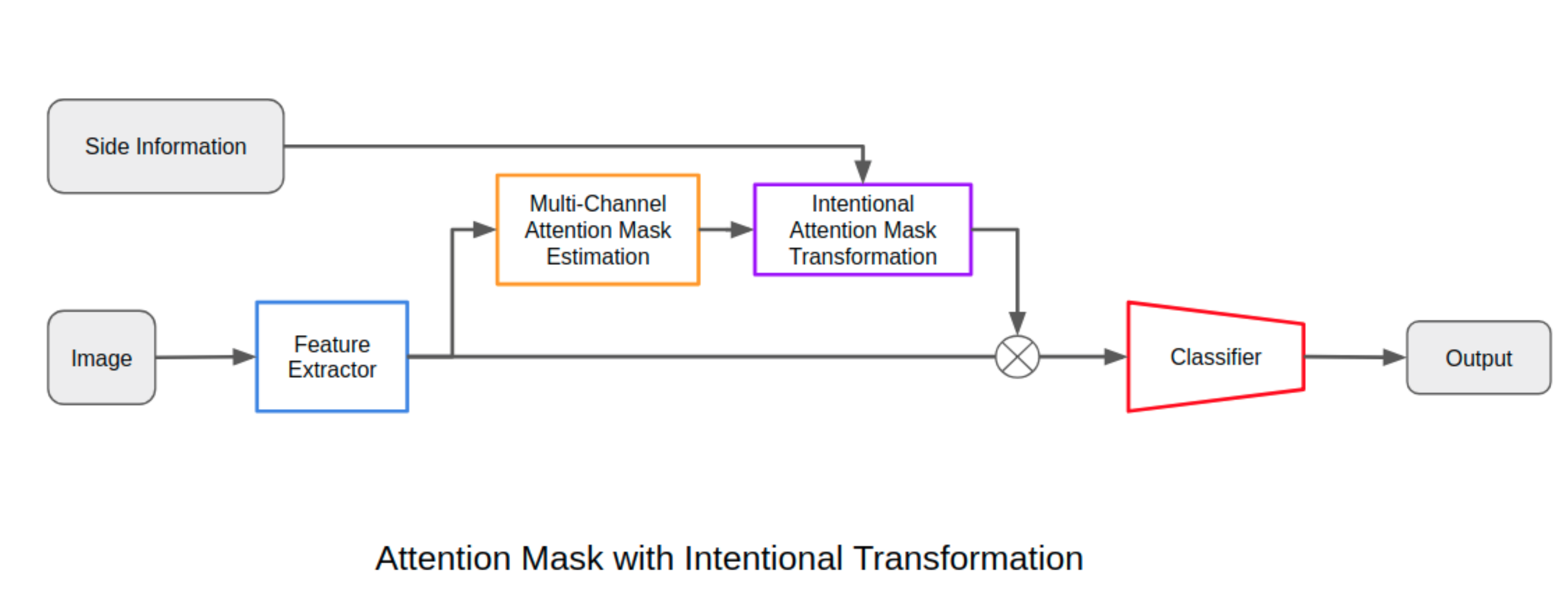}
  \caption{Overview of the intentional attention mask transformation. In this framework, robust inference can be achieved by transforming the attention mask with a transformation function that considers side information without retraining the network. In this paper, a symmetric function inspired by the tone curve is used as the transformation function, and the slope and bias of the symmetric function are used as the side information, but any function and side information (such as prior knowledge of data) can be used.}
  \label{fig:mask_transformation}
\end{figure*}

An improvement of CNNs have been usually obtained by heuristical approach or try-and-error, due to lack of interpretability of CNNs.
In contrast, we propose an intentional attention mask transformation to improve robustness of CNNs classification performance.
Our approach is to focus the important features for the classification.
The multi-channel attention sub-networks give us the information which features and which part are important for the classification.
Then, we can easily focus those features by just transforming the attention mask.
We call this operation an intentional attention mask transformation.
Figure~\ref{fig:mask_transformation} shows an overview of the intentional attention mask transformation approach. 
In our approach, we can control the property of the network by changing the intentional attention mask transformation depending on the side information.
Note that we can control the property of the network in the inference phase without re-training.

In summary, our contributions are as follows.
\begin{itemize}
    \item{We introduce an multi-channel attention sub-networks to improve interpretability of the CNN. The multi-channel attention masks show an importance of each feature to classify the attribute.}
    \item{We propose an intentional attention mask transformation to improve the robustness of CNNs against image degradation such as noise.
    The intentional attention mask transformation is performed based on the feature importance generated by the multi-channel attention sub-networks.
    }
    \item{We conduct extensive experiments on the benchmark dataset and show the usefulness of the proposed method.}
\end{itemize}

Our attention mechanism makes it possible to acquire noise-robust neural networks and provides us with high interpretability.



\section{Proposed Method}
\subsection{Network Architecture and Loss Functions}
\begin{figure*}[tb]
  \centering
  \includegraphics[scale=0.29]{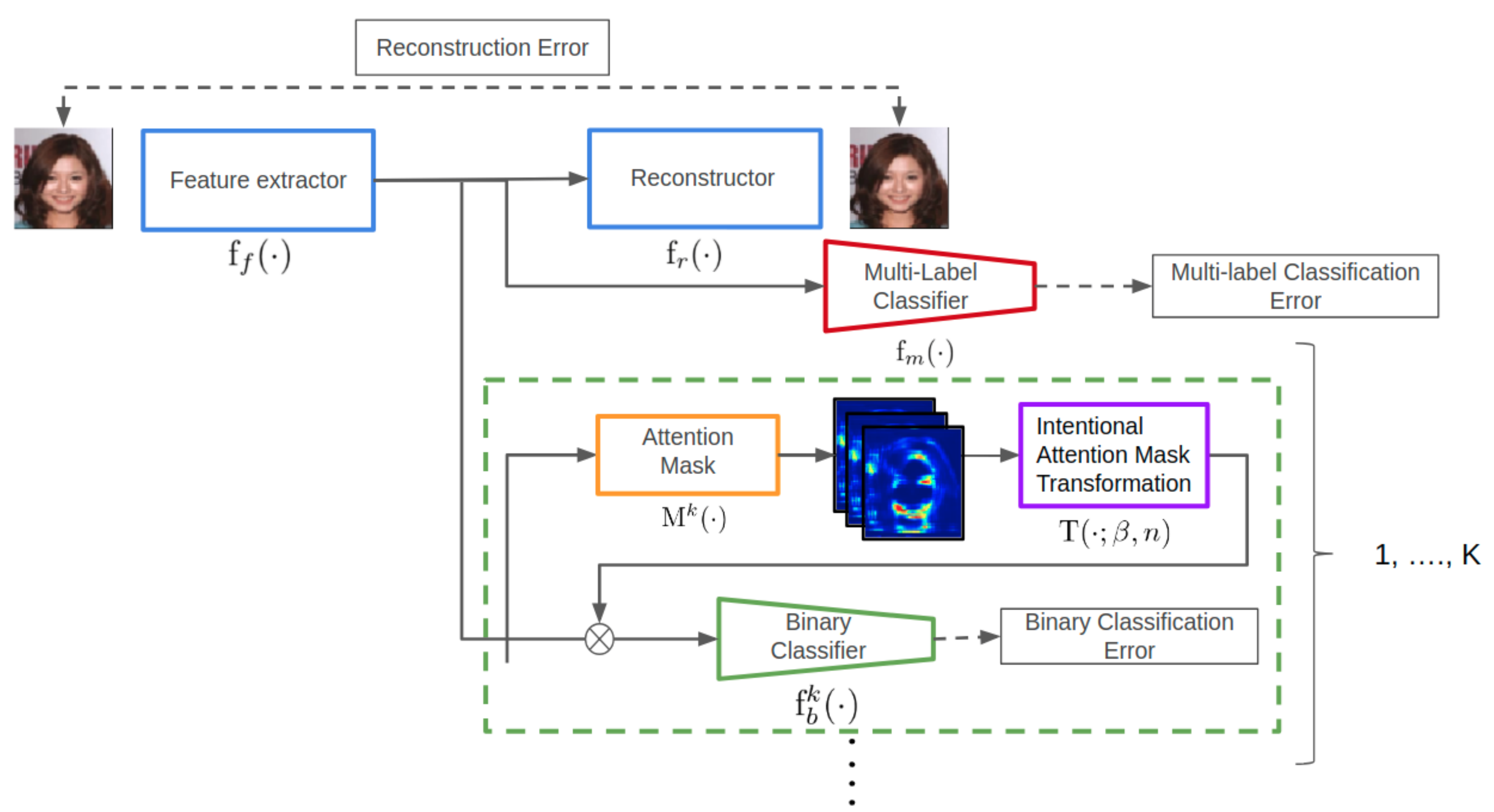}
  \caption{Overview of proposed network architecture. Here, there are $K$ binary classification components, where $K$ is the number of attributes.}
  \label{fig:network_structure1}
\end{figure*}

We aim to interpret and theorize the internal mechanisms of CNNs using attention mechanism.
Figure~\ref{fig:network_structure1} shows an overview of our proposed network architecture. In the proposed method, multiple outputs corresponding to an image with multiple attributes.

Let ${\bm X} = \{ {\bm x}_i \}^N_{i=1}$ be a sample set, ${\bm Y} = \{ {\bm y}_i \}^N_{i=1}$ be a label set, $N$ is the number of samples.
Feature extractor $f_f$ extracts generic feature used by all the following networks. This component performs feature extraction. For this component, we use the Dilation Network \cite{fisher2016multi}.

Binary classifiers $F_b = \{f_b^k \}^K_{k=1}$, where $K$ is number of attributes, are components that perform binary classification corresponding to each attribute of the image.
This network is our main component.
The loss for the binary classifiers can be expressed by a sum of binary cross entropy of each attribute as
\begin{eqnarray}
    L_{b} &=& - \frac{1}{N} \sum^N_{i=1} \sum^K_{k=1}
    \left[
    y_{i}^{k} \log{\hat{y}_{b,i}^{k}}
    + (1 - y_{i}^k)\log{(1 - \hat{y}_{b,i}^{k})} \right] \,, 
    \\
    \hat{y}_{b,i}^{k} &=& f_b^k(T(M^k(f_f({\bm x}_i))) \otimes f_f({\bm x}_i))\,,
\end{eqnarray}
where $M^k(\cdot)$ is the attention mask for $k$-th attribute, $T(\cdot)$ is the transformation function and $\otimes$ represents element-wise product operation.
Note that the attention mask $M^k$ has the same number of channels as that of feature $f_f$.
It differentiates from existing importance visualization algorithms.
For learning, we use $1 + M(\cdot)$ as the transformation function $T$. to emphasize the area of interest while keeping the low value of the attention mask following in \cite{fukui2018attention}.

Multi-label classifier is a component that classifies multiple labels.
The loss for the multi-label classifier is
\begin{eqnarray}
    L_{m} &=& - \frac{1}{N} \sum^N_{i=1} \sum^{K}_{k=1}  
    \left[ 
    y_{i}^{k} \log{\hat{y}_{m,i}^{k}}
    + (1 - y_{i}^k)\log{(1 - \hat{y}_{m,i}^{k})} \right] \,, 
    \\
    \hat{y}_{m,i}^k &=& f_m^k(f_f({\bm x}_i)) \,.
\end{eqnarray}
We put this network component to obtain better feature representation.

Reconstructor $f_r$ is a component that reconstructs a input image from extracted feature. The reconstruction loss $L_r$ is as follows.
\begin{equation}
    L_r = \frac{1}{N} \sum^N_{i=1} ({\bm x}_i - f_r(f_f({\bm x}_i)))^2 \,.
\end{equation}
This component aims to obtain better feature representation $f_f(\cdot)$.

The overall loss function is:
\begin{equation}
    L = \lambda_b \cdot L_b + \lambda_m \cdot L_m + \lambda_r \cdot L_r  + \lambda_1 \cdot \|{\bm M}\|_1\,.
\end{equation}
In the above equation, $\lambda_b$, $\lambda_m$, $\lambda_r$, and $\lambda_1$ are weight parameters of each component.
Also, $\|{\bm M}\|_1$ means L1 sparseness to the attention mask and is used to extract features that are really important for the data.
We will experimentally validate the effectiveness of the multi-label classifier and the reconstructor in section~\ref{ss:AblationStudy}.

\begin{figure*}[t]
  \centering
  \includegraphics[scale=0.35]{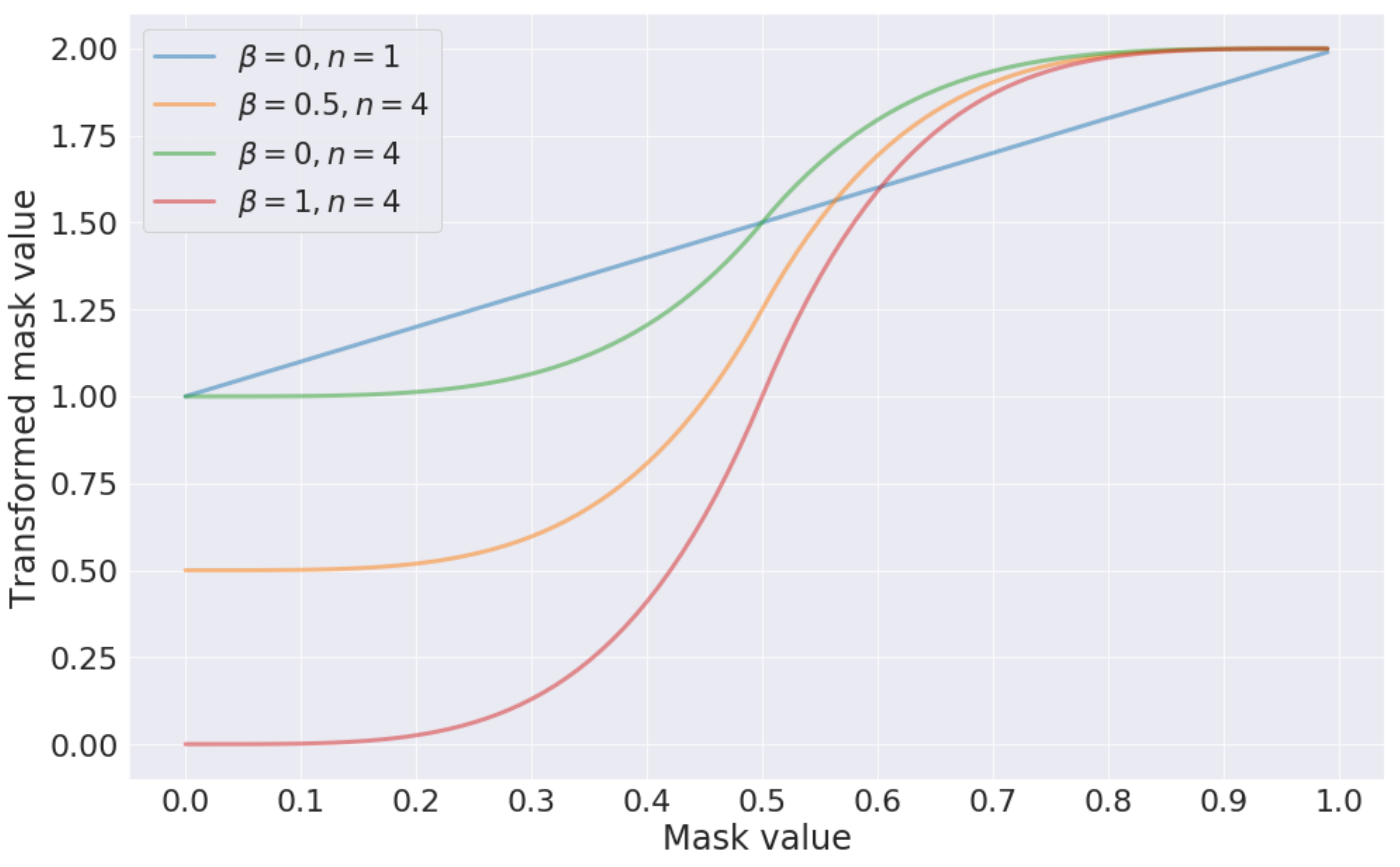}
  \caption{Visualization of transformation function. In this figure, $n$ is 0 or 4 and $\beta$ is 0 to 1. The combination of parameters for linear transformation is when $n = 1$ and $\beta = 0$.}
  \label{fig:transform_function}
\end{figure*}

\subsection{Intentional Attention Mask Transformation}
Here, we propose an intentional attention mask transformation to improve the robustness of CNNs.
Our main idea is to reduce the effect of noise by only focusing important features area for the classification of each attribute.
To achieve this goal, we introduce the following simple attention mask transformation function $g(M; n,\beta)$.
\begin{eqnarray}
    g(M^k; n,\beta) &=& (1 + \beta)\cdot h(M^k;n) - \beta, \\
    h(M^k; n) &=& \begin{cases}
    \frac{(M^k/0.5)^n}{2} & (M^k < 0.5) \\
    1 - \frac{((1-M^k)/0.5)^n}{2} & (M^k \geq 0.5)
    \end{cases}
    \,,
\end{eqnarray}
where $n$ and $\beta$ are parameters to adjust the emphasis and suppression of the mask.
The function $h(M;n)$ is symmetric with respect to 0.5, and the function $g(M; n, \beta)$ emphasizes large values in the mask $M$.
It is similar transformation as intensity tone curve in image retouching. Figure~\ref{fig:transform_function} shows the transformation function corresponding to each parameter pair.

The output of a binary classifier applying our intentional attention mask transformation is:
\begin{equation}
    \hat{y}_{b,i}^{k} = f_b^k((1+g(M^k(f_f({\bm x}_i)); n, \beta)) \otimes f_f({\bm x}_i))\,,
\end{equation}
The above equations emphasize important features about the attribute $k$ of sample $x_i$.
Emphasizing the features that are important for any given attribute makes the classifier robust to the effects of noise.

\section{Experimental Results}
We evaluate our method using the CelebA dataset \cite{liu2015deep}, which consists of 40 facial attribute labels and 202,599 images (182,637 training images and 19,962 testing images).

The parameters of the proposed method are $\lambda_b = 1$, $\lambda_m = 1$, $\lambda_r = 4$ and $\lambda_1 = 0.00001$. Also, the dimension of the attention mask and feature map is 128.

\subsection{Ablation Study and Comparisons with Existing Algorithms} \label{ss:AblationStudy}
First, we experimentally validate the effectiveness of the reconstructor and the multi-label classifier with an ablation study. In this experiment, the parameters of the transformation function are $n = 1$ and $\beta = 0$.
Table~\ref{table:ablation} shows average accuracy of the CelebA dataset.
This result demonstrates that the reconstructor and the multi-label classifier contribute to improve the performance.

We also compare the performance of our proposed network structure with existing network; MT-RBM PCA \cite{ehrlich2016facial}, LNets+ANet \cite{liu2015deep}, and FaceTracer \cite{kumar2008facetracer}.
Table~\ref{table:accuracy_celeba} shows the experimental results of the classification task in the CelebA dataset.
The proposed network achieves good performance with many attributes and all average accuracy.

\begin{figure*}[tb]
  \centering
  \includegraphics[scale=0.29]{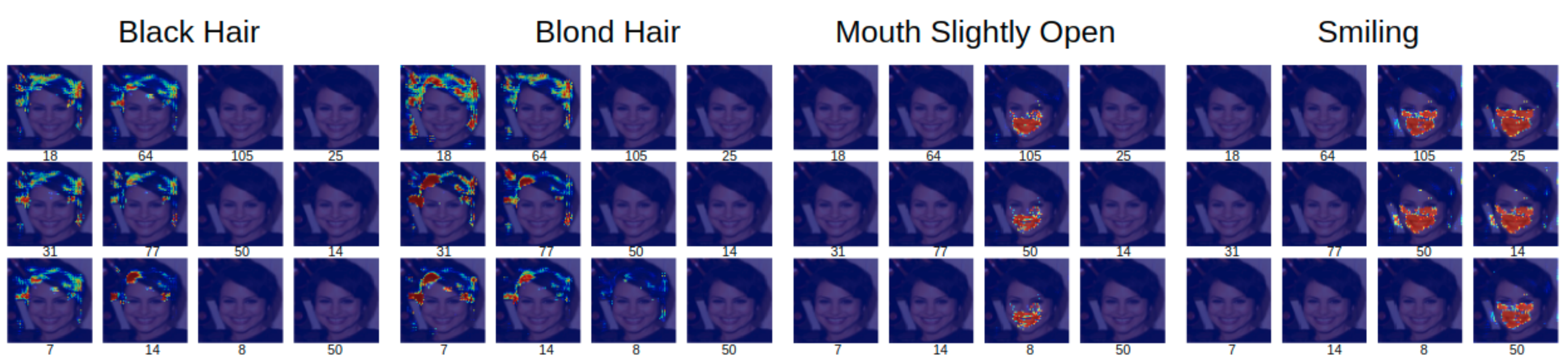}
  \caption{Visualizing attention masks on multiple facial attributes recognition. One element is one channel of the attention mask. The number under each mask means feature IDs.}
  \label{fig:attributes}
\end{figure*}

\begin{table}[t]
\centering
\caption{Ablation study for restraint networks. Comparison of classification accuracy with and without reconstructor and multi-label classifier.}
\label{table:ablation}
\begin{tabular}{l|c}
\hline
Method                                           & Average Accuracy \\ \hline
Ours                                             & {\bf 92.05}            \\
Ours w/o Reconstructor                           & 89.14            \\
Ours w/o Multi-label classifier                  & 88.12            \\
Ours w/o Reconstructor \& Multi-label classifier & 86.58            \\ \hline
\end{tabular}
\end{table}

\subsection{Visualization of the Attention Mask}
Figure~\ref{fig:attributes} shows the visualization of the attention masks by our proposed method. 
We selected several feature channels for visualization.
Each column presents top three features which has high importance for each attribute. 
Our attention masks focus on areas that may be important to attributes. 
In addition, this experimental result suggests that analysis on feature space reveals the relationship among attributes. 
For example, feature IDs 25, 14 and 50 are not used in {\it Mouse Slightly Open}, although they are used in {\it Smiling}. On the other hand, IDs 105 and 50 are used in the therapy of {\it Smiling} and {\it Mouse Slightly Open}, and ID 8 is not used in {\it Smiling}. 
Those results are consistent with human intuition, while {\it Mouse Slightly Open} should focus only on the mouth, {\it Smiling} must focus on a wide range such as eyes.
Table~\ref{table:feature_correlation_top5} lists some of the feature IDs and their highly correlated features. Our multi-channel attention mechanism makes it possible to obtain correlations among each channel of the feature map.

Table~\ref{table:correlation_top5} lists some of the attributes and their highly correlated attributes. Attributes that are intuitively similar are highly correlated. This result makes it possible to group highly correlated attributes. In addition, experimental results may even reveal potential relationships among attributes.

\begin{table*}[t]
\centering
\small
\caption{Correlation among the features. It lists the target features, highly correlated features with the target, and correlation.}
\label{table:feature_correlation_top5}
\begin{tabular}{r|r|r|r|r|r}
\hline
Target Feature & \multicolumn{5}{c}{Top5 Highly Correlated Features}                      \\ \hline
1              & 72 (0.98)  & 87 (0.95)  & 44 (0.94)  & 92 (0.94)  & 87 (0.94)  \\
32             & 111 (0.96) & 114 (0.95) & 100 (0.95) & 15 (0.94)  & 119 (0.94) \\
64             & 127 (0.97) & 15 (0.97)  & 119 (0.97) & 114 (0.97) & 57 (0.97) \\ \hline
\end{tabular}
\end{table*}

\begin{table*}[t]
\centering
\small
\caption{Correlation among the attributes. It lists the target attributes and the top five attributes that are highly correlated with the target. }
\label{table:correlation_top5}
\begin{tabular}{l|l}
\hline
Target  Attribute & Top5 Highly Correlated Attributes                                                        \\ \hline
Black Hair        & Blond Hair, Brown Hair, Bald, Wearing Hat, Gray Hair                   \\
Heavy Makeup      & Wearing Lipstick, Male, Rosy Cheeks, Attractive, Young                 \\
Bushy Eyebrows    & Bags Under Eyes, Eyeglasses, Arched Eyebrows, Heavy Makeup, Attractive \\
\hline
\end{tabular}
\end{table*}

\begin{figure*}[t]
  \centering
  \includegraphics[scale=0.35]{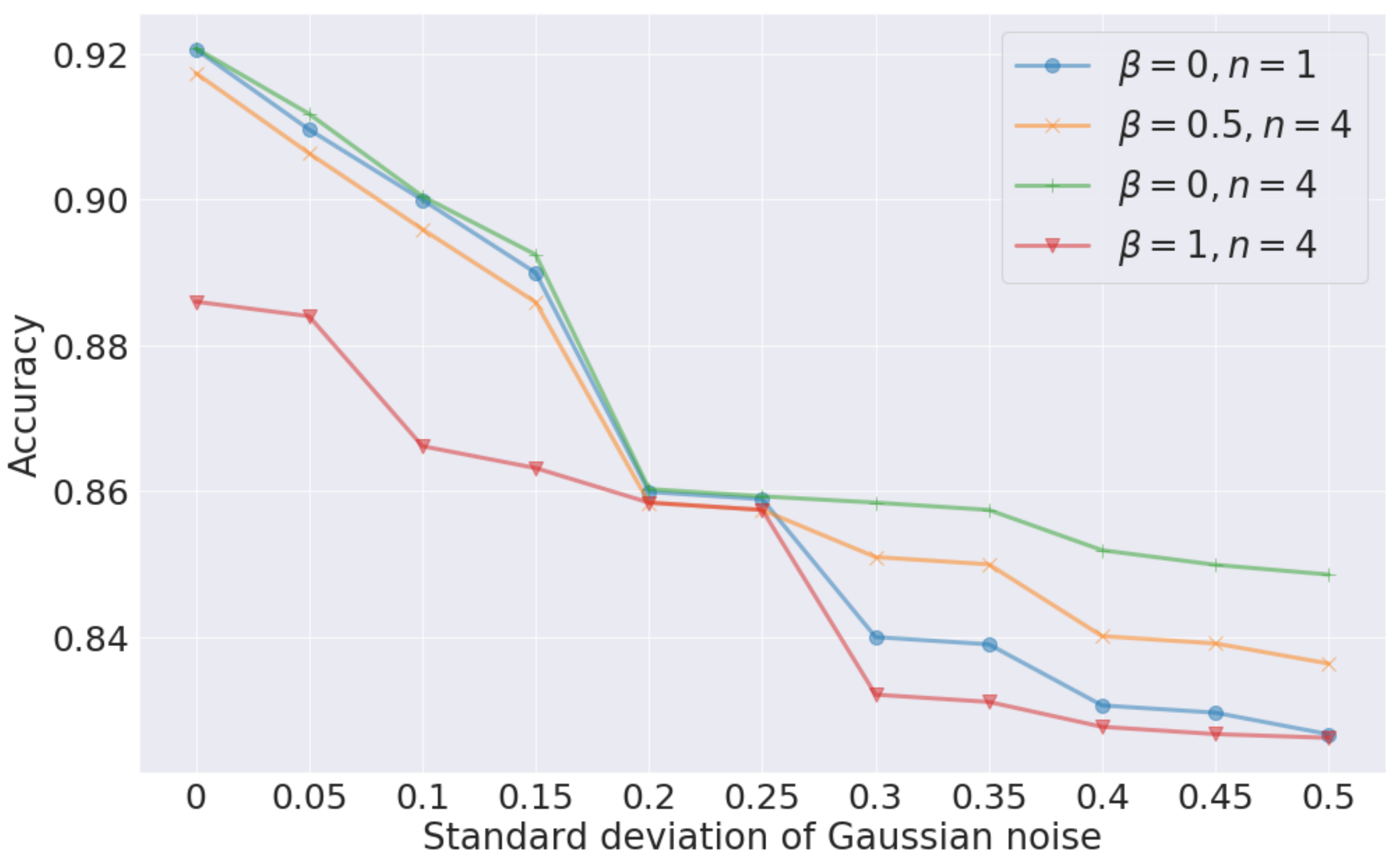}
  \caption{Transition of performance degradation of the network using the intentional attention mask transformation. We add Gaussian noise with standard deviation of 0 to 0.5 to the input image to create pseudo noisy data. Here, $n$ and $\beta$ are parameters to adjust the emphasis and suppression of the mask.}
  \label{fig:transform}
\end{figure*}

\subsection{Robustness against Noisy Inputs}
We evaluate robustness of the network by the intentional attention mask transformation.
We add Gaussian noise with standard deviation of 0 to 0.5 to the input image to create pseudo noisy data.
Figure~\ref{fig:transform} shows the experimental results of observing the transition of the performance for this noisy input data for a combination of multiple parameters.
The experimental results show that the transformation of the attention mask makes the network robust against the performance degradation due to noise. The result of $\beta = 0$ and $n = 1$ is the performance transition of the network without the intentional attention mask transformation, and by adjusting $\beta$ and $n$, the performance improvement for noisy input is achieved.

\begin{table}[t]
\centering
\vspace{0mm}
\caption{Classification accuracy on the CelebA dataset. In this experiment, MT-RBM PCA \cite{ehrlich2016facial}, LNets+ANet \cite{liu2015deep}, and FaceTracer \cite{kumar2008facetracer} are used as comparison methods.}
\label{table:accuracy_celeba}
\begin{tabular}{l|lllll}
\hline
Attribute              & Ours        & \cite{ehrlich2016facial} & \cite{liu2015deep} & \cite{kumar2008facetracer} \\ \hline
5 Shadow               & {\bf 92.85} & 90                       & 91                 & 85                 \\
Arched Eyebrows        & {\bf 81.37} & 77                       & 79                 & 76                 \\
Attractive             & 80.71       & 76                       & {\bf 81}           & 78                 \\
Bags Under Eyes        & {\bf 83.79} & 81                       & 79                 & 76                 \\
Bald                   & {\bf 98.30} & 98                       & 98                 & 89                 \\
Bangs                  & 94.10       & 88                       & {\bf 95}           & 88                 \\
Big Lips               & {\bf 70.14} & 69                       & 68                 & 64                 \\
Big Nose               & {\bf 83.67} & 81                       & 78                 & 74                 \\
Black Hair             & {\bf 88.39} & 76                       & 88                 & 70                 \\
Blond Hair             & {\bf 95.10} & 91                       & 95                 & 80                 \\
Blurry                 & {\bf 95.33} & 95                       & 84                 & 81                 \\
Brown Hair             & {\bf 86.55} & 83                       & 80                 & 60                 \\
Bushy Eyebrows         & {\bf 91.87} & 88                       & 90                 & 80                 \\
Chubby                 & {\bf 96.02} & 95                       & 91                 & 86                 \\
Double Chin            & {\bf 96.68} & 96                       & 92                 & 88                 \\
Eyeglasses             & 98.67       & 96                       & {\bf 99}           & 98                 \\
Goatee                 & {\bf 96.72} & 96                       & 95                 & 93                 \\
Gray Hair              & {\bf 97.89} & 97                       & 97                 & 90                 \\
Heavy Makeup           & 89.49       & 85                       & {\bf 90}           & 85                 \\
High Cheekbone         & 86.77       & 83                       & {\bf 87}           & 84                 \\
Male                   & 97.38       & 90                       & {\bf 98}           & 91                 \\
Mouth Open             & {\bf 93.67} & 82                       & 92                 & 87                 \\
Mustache               & 96.60       & {\bf 97}                 & 95                 & 91                 \\
Narrow Eyes            & {\bf 86.38} & 86                       & 81                 & 82                 \\
No Beard               & 94.87       & 90                       & {\bf 95}           & 90                 \\
Oval Face              & {\bf 73.33} & 73                       & 66                 & 64                 \\
Pale Skin              & {\bf 97.67} & 96                       & 91                 & 83                 \\
Pointy Nose            & {\bf 75.62} & 73                       & 72                 & 68                 \\
Recede Hair            & 93.44       & {\bf 96}                 & 89                 & 76                 \\
Rosy Cheeks            & {\bf 94.67} & 94                       & 90                 & 84                 \\
Sideburns              & {\bf 97.65} & 96                       & 96                 & 94                 \\
Smiling                & {\bf 92.28} & 88                       & 92                 & 89                 \\
Straight Hair          & {\bf 81.60} & 80                       & 73                 & 63                 \\
Wavy Hair              & {\bf 81.64} & 72                       & 80                 & 73                 \\
Earring                & {\bf 84.61} & 81                       & 82                 & 73                 \\
Hat                    & 98.92       & 97                       & {\bf 99}           & 89                 \\
Lipstick               & 92.52       & 89                       & {\bf 93}           & 89                 \\
Necklace               & 86.37       & {\bf 87}                 & 71                 & 68                 \\
Necktie                & {\bf 96.30} & 94                       & 93                 & 86                 \\
Young                  & {\bf 87.00} & 81                       & {\bf 87}           & 80                 \\ \hline
Average                & {\bf 92.05} & 87                       & 87                 & 81                 \\ \hline
\end{tabular}
\end{table}

\section{Conclusion and Discussion}
We proposed a novel network architecture and attention mechanism that can give a visual explanation of CNNs. 
Our multi-channel attention mechanism makes it possible to obtain correlations among each channel of the feature map. We suggest that analysis of feature maps obtained by the proposed method is highly versatile and lead to a broad range of applied research, such as improvement of classification accuracy, network pruning, image generation, and other applications.
As one of such applications, we have shown that intentional transformation of the attention mask can improve the robustness of CNNs.

\end{document}